\documentclass[runningheads]{llncs}
\usepackage{dsfont}
\usepackage{amsmath}
\usepackage{amssymb}
\usepackage{graphicx}
\usepackage{subfigure}
\usepackage{wrapfig}
\usepackage{cite}
\usepackage{bm}
\usepackage{hyperref}
\usepackage{multirow}
\usepackage{tabularx}
\usepackage{xcolor}

\begin{document}
	
	\mainmatter  
	
	\title{Knowledge Transfer for Few-shot Segmentation of Novel White Matter Tracts}
	
	
	%
	%
	
	\author{Qi Lu\orcidID{0000-0003-4659-1443} \and
		Chuyang Ye\orcidID{0000-0001-5839-1559}}

	\authorrunning{Lu et al.}
	
	\titlerunning{Few-shot Segmentation of Novel WM Tracts}
	
	\institute{School of Information and Electronics, Beijing Institute of Technology, Beijing, China\\
		\email{luqily@163.com} \quad\quad\quad \email{chuyang.ye@bit.edu.cn}\\}
	
	%
	%
	
	\toctitle{}
	\tocauthor{}
	\maketitle
	\begin{abstract}
		\textit{Convolutional neural networks}~(CNNs) have achieved state-of-the-art performance for \textit{white matter}~(WM) tract segmentation based on \textit{diffusion magnetic resonance imaging}~(dMRI). These CNNs require a large number of manual delineations of the WM tracts of interest for training, which are generally labor-intensive and costly. The expensive manual delineation can be a particular disadvantage when novel WM tracts, i.e., tracts that have not been included in existing manual delineations, are to be analyzed. To accurately segment novel WM tracts, it is desirable to transfer the knowledge learned about existing WM tracts, so that even with only a few delineations of the novel WM tracts, CNNs can learn adequately for the segmentation. In this paper, we explore the transfer of such knowledge to the segmentation of novel WM tracts in the few-shot setting. Although a classic fine-tuning strategy can be used for the purpose, the information in the last task-specific layer for segmenting existing WM tracts is completely discarded. We hypothesize that the weights of this last layer can bear valuable information for segmenting the novel WM tracts and thus completely discarding the information is not optimal. In particular, we assume that the novel WM tracts can correlate with existing WM tracts and the segmentation of novel WM tracts can be predicted with the logits of existing WM tracts. In this way, better initialization of the last layer than random initialization can be achieved for fine-tuning. Further, we show that a more adaptive use of the knowledge in the last layer for segmenting existing WM tracts can be conveniently achieved by simply inserting a warmup stage before classic fine-tuning. The proposed method was evaluated on a publicly available dMRI dataset, where we demonstrate the benefit of our method for few-shot segmentation of novel WM tracts.

	\end{abstract}
	
	\section{Introduction}
	\label{sec:intro}
	
	\textit{White matter}~(WM) tract segmentation based on~\textit{diffusion magnetic resonance imaging}~(dMRI) allows identification of specific WM pathways~\cite{wasserthal2018tractseg}, which are linked to brain development, function, and disease~\cite{zhang2020deep}. 
	To achieve automated and accurate WM tract segmentation, \textit{convolutional neural networks}~(CNNs) have been applied to the segmentation task, and they have achieved state-of-the-art performance. For example, CNNs can be used to label fiber streamlines, which represent WM pathways and are computed with tractography~\cite{Jeurissen2017Diffusion}, based on the feature maps extracted for each fiber streamline~\cite{zhang2020deep}. The labeled streamlines form the representation of specific WM tracts. It is also possible to perform volumetric WM tract segmentation that directly labels each voxel in a dMRI scan according to the WM tracts it belongs to based on diffusion feature maps~\cite{li2020neuro4neuro} or fiber orientation maps~\cite{wasserthal2018tractseg, lu2020white}. Since volumetric segmentation assigns labels to voxels instead of fiber streamlines, the tractography step is not necessarily needed for this type of methods.
	
	CNN-based methods of WM tract segmentation generally require a large number of manual delineations of WM tracts for training. These delineations can be very labor-intensive and costly. Although there can be existing delineations of certain WM tracts that are accumulated throughout time, the expensive delineation can be a particular disadvantage when novel WM tracts---i.e., WM tracts that have not been included in existing manual delineations---are to be analyzed. It is desirable to perform accurate segmentation of these novel WM tracts with only a few delineations, and this can be achieved by exploiting the knowledge learned about existing WM tracts. For example, the classic fine-tuning strategy~\cite{tajbakhsh2016convolutional} can be used for this few-shot setting, where knowledge about existing WM tracts is transferred to the novel WM tracts by replacing the last layer of the network for segmenting existing WM tracts with a randomly initialized output layer for the novel WM tracts. Then, all network parameters are jointly learned from the limited number of annotations of novel WM tracts. 
	
	Although classic fine-tuning can be effective for few-shot segmentation of novel WM tracts, it completely discards the information in the last task-specific layer for segmenting existing WM tracts. Since different WM tracts can be correlated, the discarded layer may bear valuable information that is relevant to the novel WM tracts, and thus classic fine-tuning may be suboptimal. Therefore, in this paper, we further explore the transfer of knowledge learned from abundant annotations of existing WM tracts to the segmentation of novel WM tracts in the few-shot setting. In particular, we focus on the scenario where only the model trained for segmenting existing WM tracts is available and the training data for existing WM tracts is inaccessible. This scenario is common for medical imaging due to privacy or other practical concerns~\cite{burton2015data}. Also, this work focuses on methods of volumetric segmentation because they do not require the step of tractography that could be sensitive to the choice of algorithms and hyperparameters, although the proposed idea may be extended to those methods based on fiber streamlines as well. 
	
	We assume that knowledge about segmenting existing WM tracts can inform the segmentation of novel WM tracts, and this can be achieved from the logits---the unnormalized predictions before the final activation function---of existing WM tracts. For simplicity, a logistic regression model is used for the prediction, which is then combined with the last layer for segmenting existing WM tracts to provide better network initialization for segmenting the novel WM tracts. In this way, all knowledge learned for segmenting existing WM tracts, including the information in the last layer, is transferred to the segmentation of novel WM tracts. Further, we show that this problem formulation motivates a more adaptive transfer of the knowledge, and it turns out that this adaptive knowledge transfer is simply equivalent to the insertion of a warmup stage before classic fine-tuning. We evaluated the proposed method using the publicly available \textit{Human Connectome Project}~(HCP) dataset~\cite{van2013wu}. Experimental results show that our method improves the performance of few-shot segmentation of novel WM tracts.

	\section{Methods}
	\label{sec:method}
	\subsection{Problem Formulation and Classic Fine-tuning}
	
	Suppose we have a CNN-based segmentation model trained with abundant annotations for a set of WM tracts, and the annotations may not be accessible. We are interested in the segmentation of a novel set of WM tracts that are not considered during the training of the given model. Only a few manual annotations are available for these novel WM tracts, and our goal is to achieve decent segmentation performance for the novel WM tracts given the scarce annotations. 
	To achieve such a goal, a common practice is to transfer the knowledge in the model learned for segmenting existing WM tracts to the segmentation of novel WM tracts. 
	Intuitively, the knowledge transfer can be performed with the classic fine-tuning strategy~\cite{tajbakhsh2016convolutional}, which we formulate mathematically as follows.  
	
	For convenience, we denote the network models for segmenting existing and novel WM tracts by $\mathcal{M}_{\mathrm{e}}$ and $\mathcal{M}_{\mathrm{n}}$, respectively. In classic fine-tuning, $\mathcal{M}_{\mathrm{e}}$ and $\mathcal{M}_{\mathrm{n}}$ share the same network structure except for the last layer, which is task-specific.
	Suppose the input image is $\mathbf{X}$, the task-specific weights in the last layer $L_\mathrm{e}$ of $\mathcal{M}_{\mathrm{e}}$ and the last layer $L_\mathrm{n}$ of $\mathcal{M}_{\mathrm{n}}$ are denoted by $\bm{\theta}_{\mathrm{e}}$ and $\bm{\theta}_{\mathrm{n}}$, respectively, and the other weights in $\mathcal{M}_{\mathrm{e}}$ or $\mathcal{M}_{\mathrm{n}}$ are denoted by $\bm{\theta}$. 
	From $\mathbf{X}$ a multi-channel feature map $\mathbf{F}$ is computed with a mapping $f(\mathbf{X}; \bm{\theta})$ parameterized by $\bm{\theta}$:
	\begin{equation}
	\mathbf{F} = f(\mathbf{X}; \bm{\theta}),
	\end{equation} 
	and the segmentation probability map $\mathbf{P}_{\mathrm{e}}$ or $\mathbf{P}_{\mathrm{n}}$ for existing or novel WM tracts is computed from $\mathbf{F}$ with $L_\mathrm{e}$ or $L_\mathrm{n}$ using another mapping $g_{\mathrm{e}}(\mathbf{F}; \bm{\theta}_{\mathrm{e}})$ or $g_{\mathrm{n}}(\mathbf{F}; \bm{\theta}_{\mathrm{n}})$ parameterized by $\bm{\theta}_{\mathrm{e}}$ or $\bm{\theta}_{\mathrm{n}}$, respectively:
	\begin{eqnarray}
	\mathbf{P}_{\mathrm{e}} = g_{\mathrm{e}}(\mathbf{F}; \bm{\theta}_{\mathrm{e}})=g_{\mathrm{e}}(f(\mathbf{X}; \bm{\theta}); \bm{\theta}_{\mathrm{e}}) &\rm{and}&
	\mathbf{P}_{\mathrm{n}} = g_{\mathrm{n}}(\mathbf{F}; \bm{\theta}_{\mathrm{n}})=g_{\mathrm{n}}(f(\mathbf{X}; \bm{\theta}); \bm{\theta}_{\mathrm{n}}).
	\end{eqnarray}
	
	Instead of directly training $\mathcal{M}_{\mathrm{n}}$ from scratch---i.e., $\bm{\theta}$ and $\bm{\theta}_{\mathrm{n}}$ are randomly initialized---using the scarce annotations of novel WM tracts, in classic fine-tuning the information in $\mathcal{M}_{\mathrm{e}}$ is exploited.
	Because $\mathcal{M}_{\mathrm{e}}$ is trained by minimizing the difference between $\mathbf{P}_{\mathrm{e}}$ and the abundant annotations of existing WM tracts, the learned values $\tilde{\bm{\theta}}$ of the weights $\bm{\theta}$ for segmenting existing WM tracts can provide useful information about feature extraction. Thus, $\tilde{\bm{\theta}}$ is used to initialize $\bm{\theta}$ for training $\mathcal{M}_{\mathrm{n}}$, and only $\bm{\theta}_{\mathrm{n}}$ is randomly initialized. In this way, the knowledge learned for segmenting existing WM tracts can be transferred to the segmentation of novel WM tracts, and this classic fine-tuning strategy has been proved successful in a variety of image processing applications~\cite{tajbakhsh2016convolutional}.
	
	\subsection{Knowledge Transfer for Few-shot Segmentation of Novel WM Tracts}
	\label{sec:kt}
	
	Although the classic fine-tuning strategy can be used for the few-shot segmentation of novel WM tracts, it completely discards the information about $\bm{\theta}_{\mathrm{e}}$ in the last layer $L_{\mathrm{e}}$ learned for existing WM tracts. 
	We hypothesize that the information in these discarded weights could also bear information relevant to the segmentation of novel WM tracts. For example, in a considerable number of voxels, WM tracts are known to co-occur as crossing fiber tracts~\cite{ginsburger2019medusa}. 
	Thus, it is reasonable to assume that existing and novel WM tracts can be correlated and novel WM tracts could be predicted from existing WM tracts, and this assumption allows us to explore the discarded information in $L_{\mathrm{e}}$ as well for training $\mathcal{M}_{\mathrm{n}}$.
	
	Suppose $\mathbf{P}_{\mathrm{e}}^{v}$ and $\mathbf{P}_{\mathrm{n}}^{v}$ are the vectors of segmentation probabilities at the $v$-th voxel of $\mathbf{P}_{\mathrm{e}}$ and $\mathbf{P}_{\mathrm{n}}$, respectively, where $v\in\{1,2,\ldots,V\}$ and $V$ is the total number of voxels.
	In existing segmentation networks, $L_\mathrm{e}$ and $L_\mathrm{n}$ generally use a convolution with a kernel size of one to classify each voxel (e.g., see~\cite{wasserthal2018tractseg}), which is equivalent to matrix multiplication (plus a bias vector) at each voxel. 
	Therefore, we rewrite the task-specific weights as $\bm{\theta}_{\mathrm{e}} = \{\mathbf{W}_{\mathrm{e}},\bm{b}_{\mathrm{e}}\}$ and $\bm{\theta}_{\mathrm{n}} = \{\mathbf{W}_{\mathrm{n}},\bm{b}_{\mathrm{n}}\}$, so that the segmentation probabilities can be explicitly expressed as
	\begin{eqnarray}
	\mathbf{P}_{\mathrm{e}}^{v} = \sigma \left( \mathbf{W}_{\mathrm{e}} \mathbf{F}^{v} + \bm{b}_{\mathrm{e}} \right) &\rm{and}&
	\mathbf{P}_{\mathrm{n}}^{v} = \sigma \left( \mathbf{W}_{\mathrm{n}} \mathbf{F}^{v} + \bm{b}_{\mathrm{n}} \right),
	\label{eq:prob1}
	\end{eqnarray}
	where $\mathbf{F}^{v}$ represents the feature vector at the $v$-th voxel of the feature map~$\mathbf{F}$,  and $\sigma(\cdot)$ is the sigmoid activation because there can be multiple WM tracts in a single voxel.
	
	In classic fine-tuning the information about $\mathbf{W}_{\mathrm{e}}$ and $\bm{b}_{\mathrm{e}}$ is completely discarded. However, according to our assumption, it is possible to exploit $\mathbf{W}_{\mathrm{e}}$ and $\bm{b}_{\mathrm{e}}$ to provide better initialization for $\mathbf{W}_{\mathrm{n}}$ and $\bm{b}_{\mathrm{n}}$. 
	To this end, we investigate the prediction of novel WM tracts with the logits $\mathbf{H}_{\mathrm{e}}$ of existing WM tracts given by the trained $\mathcal{M}_{\mathrm{e}}$. 
	For simplicity, this prediction is achieved with a logistic regression.
	We denote the logit vector at voxel $v$ given by the trained $\mathcal{M}_{\mathrm{e}}$ by $\mathbf{H}^{v}_{\mathrm{e}}=(h_{\mathrm{e},1}^{v},\ldots,h_{\mathrm{e},M}^{v})^\mathsf{T}$, where $M$ is the number of existing WM tracts. 
	Then, the prediction $p_{\mathrm{e}\rightarrow\mathrm{n},j}^{v}$ of the $j$-th novel WM tract at voxel $v$ from the information of existing WM tracts is given by
	\begin{eqnarray}
	p_{\mathrm{e}\rightarrow\mathrm{n},j}^{v} = \frac{1}{1 +  \exp\left( -(b_{j} + \sum_{i=1}^{M} w_{ij}h^{v}_{\mathrm{e},i})\right)},
	\end{eqnarray}
	where $w_{ij}$ and $b_{j}$ are the regression parameters to be determined.
	
	Suppose the total number of novel WM tracts is $N$. Combining the prediction of all novel WM tracts into $\mathbf{P}^{v}_{\mathrm{e}\rightarrow\mathrm{n}}$, we simply have
	\begin{eqnarray}
	\mathbf{P}^{v}_{\mathrm{e}\rightarrow\mathrm{n}} = \sigma \left( \mathbf{W} \mathbf{H}^{v}_{\mathrm{e}} + \bm{b} \right),
	\end{eqnarray}
	where  
	\begin{eqnarray}
	\mathbf{W}= 
	\begin{bmatrix}  
	w_{11} & \dots & w_{1M}\\
	\vdots  & \ddots & \vdots \\
	w_{N1} & \dots & w_{NM}
	\end{bmatrix}
	\mathrm{and} \,\,
	\bm{b}= [b_{1},\ldots,b_{N}]^{\mathsf{T}}.
	\end{eqnarray}
	Note that $\mathbf{H}^{v}_{\mathrm{e}}=\widetilde{\mathbf{W}}_{\mathrm{e}} \widetilde{\mathbf{F}}^{v} + \tilde{\bm{b}}_{\mathrm{e}}$, where $\widetilde{\mathbf{F}}^{v}$ corresponds to the $v$-th voxel of $\widetilde{\mathbf{F}}=f(\mathbf{X}; \tilde{\bm{\theta}})$ that is computed with the weights $\tilde{\bm{\theta}}$ learned for existing WM tracts, and $\widetilde{\mathbf{W}}_{\mathrm{e}}$ and $\tilde{\bm{b}}_{\mathrm{e}}$ are the values of $\mathbf{W}_{\mathrm{e}}$ and $\bm{b}_{\mathrm{e}}$ learned for existing WM tracts, respectively. Then, we have 
	\begin{eqnarray}
	\mathbf{P}^{v}_{\mathrm{e}\rightarrow\mathrm{n}} = \sigma \left( \mathbf{W}  \left(\widetilde{\mathbf{W}}_{\mathrm{e}} \widetilde{\mathbf{F}}^{v} + \tilde{\bm{b}}_{\mathrm{e}}\right) + \bm{b} \right) = \sigma \left( \mathbf{W} \widetilde{\mathbf{W}}_{\mathrm{e}} \widetilde{\mathbf{F}}^{v} + \mathbf{W} \tilde{\bm{b}}_{\mathrm{e}} + \bm{b} \right).
	\label{eq:prob2}
	\end{eqnarray}
	
	Comparing Eqs.~(\ref{eq:prob1}) and (\ref{eq:prob2}), we notice that instead of being randomly initialized, $\bm{\theta}_{\mathrm{n}}$ may be better initialized using the information in $\bm{\theta}_{\mathrm{e}}=\{\mathbf{W}_{\mathrm{e}},\bm{b}_{\mathrm{e}}\}$. Here, $\mathbf{W}$ and $\bm{b}$ still need to be computed for initializing $\bm{\theta}_{\mathrm{n}}$, and they can be computed by minimizing the difference between $\mathbf{P}^{v}_{\mathrm{e}\rightarrow\mathrm{n}}$ and the annotation of novel WM tracts. Note that although there are only a few annotations of novel WM tracts, they are sufficient for the computation of $\mathbf{W}$ and $\bm{b}$ because the number of unknown parameters is also drastically reduced.
	Then, suppose the estimates of $\mathbf{W}$ and $\bm{b}$ are $\widetilde{\mathbf{W}}$ and $\tilde{\bm{b}}$, respectively; $\mathbf{W}_{\mathrm{n}}$ and $\bm{b}_{\mathrm{n}}$ are initialized as
	\begin{eqnarray}
	\mathbf{W}_{\mathrm{n}} \leftarrow  \widetilde{\mathbf{W}} \widetilde{\mathbf{W}}_{\mathrm{e}}  &\rm{and}&
	\bm{b}_{\mathrm{n}} \leftarrow  \widetilde{\mathbf{W}} \tilde{\bm{b}}_{\mathrm{e}} + \tilde{\bm{b}}.
	\label{eq:decomp2}
	\end{eqnarray}
	Finally, with $\bm{\theta}$ initialized by $\tilde{\bm{\theta}}$ like in classic fine-tuning, all network weights are learned jointly for $\mathcal{M}_{\mathrm{n}}$ using the scarce annotations of novel WM tracts.
	
	\subsection{A Better Implementation with Warmup}
	\label{sec:warm}
	The derivation above suggests a possible way of using all information in $\mathcal{M}_{\mathrm{e}}$. However, it is possible to have a more convenient implementation. If we let $\mathbf{W}'=\mathbf{W} \widetilde{\mathbf{W}}_{\mathrm{e}}$ and $\bm{b}'=\mathbf{W} \tilde{\bm{b}}_{\mathrm{e}} + \bm{b}$, Eq.~(\ref{eq:prob2}) becomes
	\begin{eqnarray}
	\mathbf{P}^{v}_{\mathrm{e}\rightarrow\mathrm{n}} &=& \sigma \left( \mathbf{W}' \widetilde{\mathbf{F}}^{v} + \bm{b}' \right).
	\label{eq:prob3}
	\end{eqnarray}
	This suggests that we can directly estimate $\mathbf{W}'$ and $\bm{b}'$ and use the estimated values to initialize $\bm{\theta}_{\mathrm{n}}$. This is equivalent to inserting a warmup stage before the classic fine-tuning, and the information in $\bm{\theta}_{\mathrm{e}}$ becomes redundant with such a fine-tuning strategy (but not with classic fine-tuning). Specifically, given the trained model $\mathcal{M}_{\mathrm{e}}$, for $\mathcal{M}_{\mathrm{n}}$ we first set $\bm{\theta}\leftarrow\tilde{\bm{\theta}}$, fix $\bm{\theta}$, and learn $\bm{\theta}_{\mathrm{n}}$ (randomly initialized) from the scarce annotations of novel WM tracts. With the values of $\bm{\theta}_{\mathrm{n}}$ learned in the first stage, we then jointly fine-tune the weights $\bm{\theta}$ and $\bm{\theta}_{\mathrm{n}}$ using the annotated novel WM tracts.
	
	This implementation not only is more convenient than the derivation in Sect.~\ref{sec:kt}, but also could lead to better performance for the following reasons. First, the warmup strategy is not restricted to the decomposition in Eq.~(\ref{eq:decomp2}) and allows a more adaptive use of the information in $\bm{\theta}_{\mathrm{e}}$. It can find the initialization corresponding to the decomposition as well as possibly better initialization that may not be decomposed as Eq.~(\ref{eq:decomp2}). Second, even for the case where the decomposed form allows the best initialization, the separate computation of $\{\widetilde{\mathbf{W}}, \tilde{\bm{b}}\}$ and $\{\widetilde{\mathbf{W}}_{\mathrm{e}}, \tilde{\bm{b}}_{\mathrm{e}}\}$ could accumulate the error of each computation and slightly degrade the initialization, whereas directly estimating $\mathbf{W}'$ and $\bm{b}'$ avoids the problem.

	\subsection{Implementation Details}
	We use the state-of-the-art TractSeg architecture proposed in~\cite{wasserthal2018tractseg} as our backbone network.\footnote{Our method can also be integrated with other networks for volumetric WM tract segmentation.} 
	TractSeg is inspired by the U-net architecture~\cite{ronneberger2015u}, and it performs 2D processing for each orientation separately.
	For test scans, the results of each orientation are fused by averaging for the final 3D WM tract segmentation.
	
	In TractSeg, the network inputs are fiber orientation maps computed with multi-shell multi-tissue constrained spherical deconvolution~\cite{jeurissen2014multi}. 
	A maximum number of three fiber orientations is used at each voxel, and thus the input has nine channels. For voxels with fewer than three fiber orientations, the values in the corresponding empty channels are set to zero. The outputs of TractSeg are probability maps of WM tracts.
	
	We have implemented the proposed method using PyTorch~\cite{paszke2019pytorch}.
	Like~\cite{wasserthal2018tractseg}, for all network training the cross-entropy loss is minimized using Adamax~\cite{kingma2014adam} with a learning rate of 0.001 and a batch size of 47~\cite{wasserthal2019combined}; we also use dropout~\cite{srivastava2014dropout} with a probability of 0.4~\cite{wasserthal2018tractseg}. 200 training epochs are used to ensure training convergence. Model selection is performed according to the epoch with the best Dice score on a validation set.

	\section{Results}
	\label{sec:exp}
	
	\begin{table}[!t]
		\caption{A list of the 12 novel WM tracts and their abbreviations.}
		\label{tab:tract}
		\centering
		\resizebox{0.99\columnwidth}{!}{
			\begin{tabular}{l l l |l l l }
				\\
				\hline
				\hline	
				& WM tract name & abbreviation & & WM tract name & abbreviation\\	
				\hline
				1 & Corticospinal tract left & CST\_left  &
				7 & Optic radiation left & OR\_left  \\
				2 & Corticospinal tract right & CST\_right   &
				8 & Optic radiation right & OR\_right \\
				3 & Fronto-pontine tract left & FPT\_left   &  			
				9 & Inferior longitudinal fascicle left & ILF\_left \\
				4 & Fronto-pontine tract right & FPT\_right   & 
				10 & Inferior longitudinal fascicle right & ILF\_right \\
				5 & Parieto-occipital pontine left & POPT\_left  &  			
				11 & Uncinate fascicle left & UF\_left \\
				6 & Parieto-occipital pontine right & POPT\_right  &  			
				12 & Uncinate fascicle right & UF\_right \\
				\hline			
				\hline
				
			\end{tabular}
		}
	\end{table}
	
	\begin{figure}[!t]
		\centering
		\includegraphics[width=0.9\columnwidth]{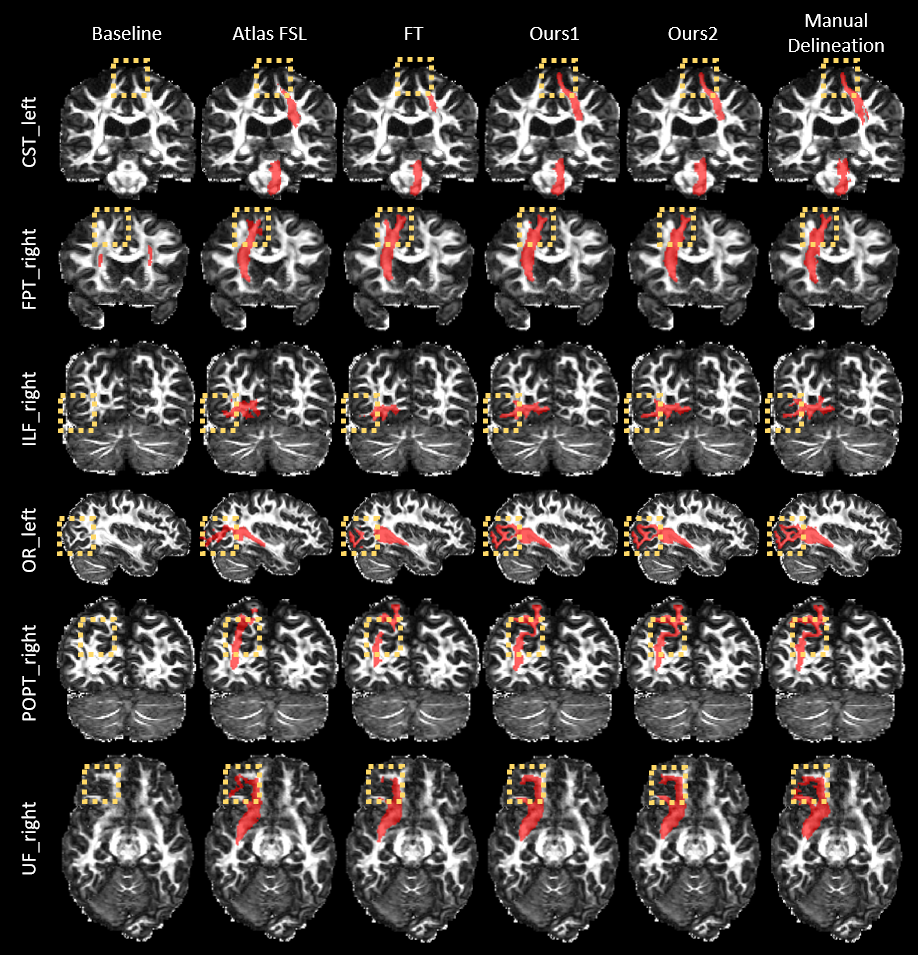}
		\caption{Cross-sectional views of the segmentation results (red) overlaid on the fractional anisotropy maps for representative test subjects and novel WM tracts. The manual delineations are also shown for reference. Note the highlighted regions for comparison.}
		\label{fig:qualitative}
	\end{figure}
	
	\begin{figure}[!t]
		\centering
		\includegraphics[width=0.81\columnwidth]{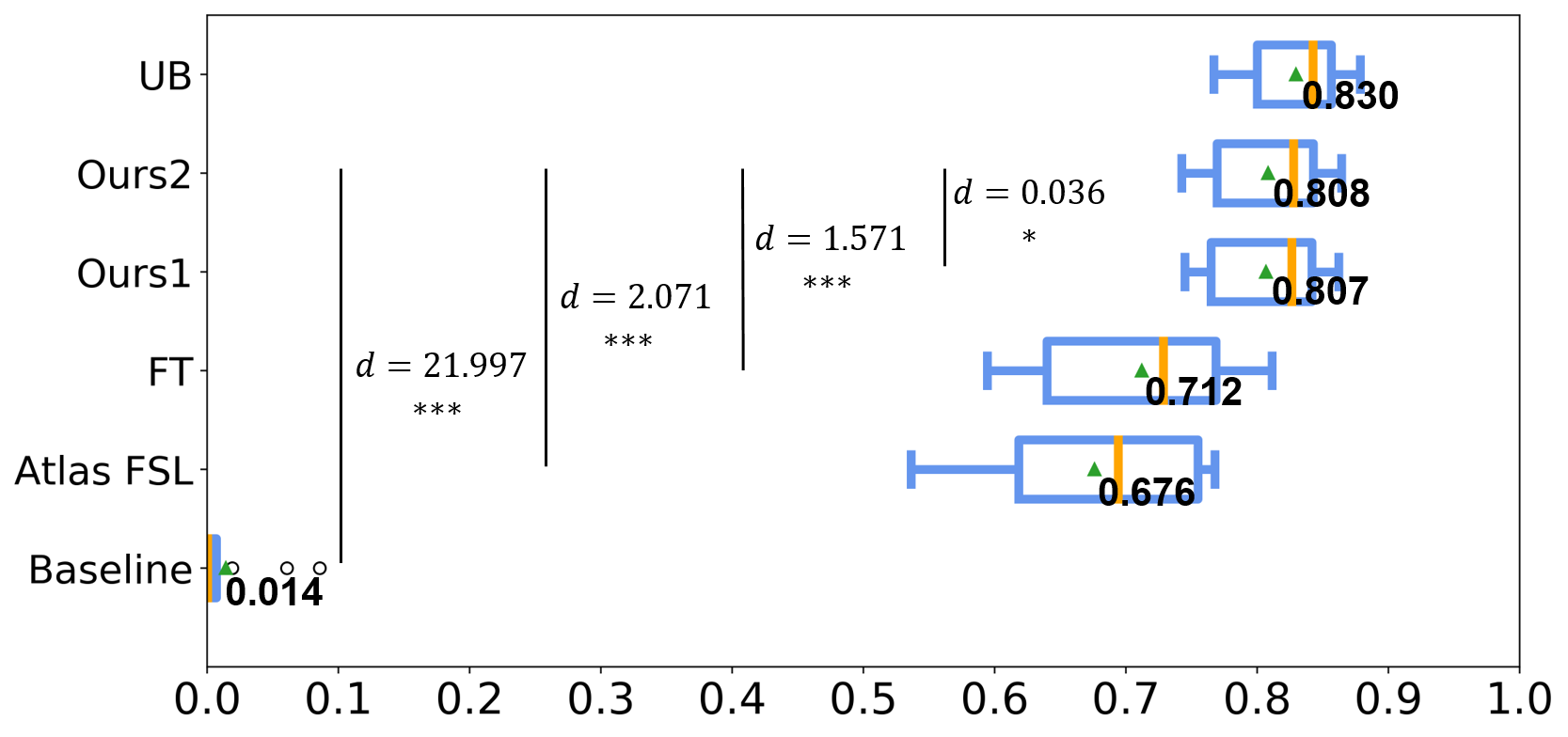}
		\caption{Boxplots of the average Dice coefficient for each tract. The means of the average Dice coefficients are indicated. The effect sizes (Cohen's $d$) between Ours2 and the other methods are also listed. Asterisks indicate that the difference between Ours2 and the other method is significant using a paired Student's $t$-test. ($^{*}p<0.05$, $^{***}p<0.001$.)
		}
		\label{fig:boxplot_mean}
	\end{figure}
	
	\subsection{Data Description and Experimental Settings}
	\label{sec:data}
	
	We used the preprocessed dMRI scans in the HCP dataset~\cite{glasser2013minimal,van2013wu} for evaluation, which were acquired with 270 diffusion gradients ($b=1000$, $2000$, and $3000~\mathrm{s}/\mathrm{mm}^2$) and 18 $b0$ images~\cite{sotiropoulos2013advances}. The image resolution is 1.25 $\mathrm{mm}$ isotropic. 
	
	We selected 60 and 12 tracts as the existing and novel WM tracts, respectively, i.e., $M=60$ and $N=12$. These 72 WM tracts in total are the same WM tracts considered in~\cite{wasserthal2018tractseg}. The list of the novel WM tracts is shown in Table~\ref{tab:tract}, and they were randomly selected from the bilateral WM tracts. The existing WM tracts correspond to the remaining WM tracts in~\cite{wasserthal2018tractseg}.\footnote{Refer to~\cite{wasserthal2018tractseg} for the list of these remaining WM tracts.} 
	A segmentation model was trained for the existing WM tracts using 65 dMRI scans, which were split into a training set comprising 52 dMRI scans and a validation set comprising 13 dMRI scans. For segmenting the novel WM tracts, we selected four other dMRI scans for network fine-tuning, where three dMRI scans were used as the training set and one dMRI scan was used as the validation set. For evaluation, the proposed method was applied to 30 test scans that were different from all the training and validation scans described above. 
	The annotations of all training, validation, and test scans are provided by~\cite{wasserthal2018tractseg}.
	
	\subsection{Evaluation of Segmentation Accuracy}
	\label{sec:accuracy}
	We first evaluated the accuracy of the proposed method, where the segmentation model for existing WM tracts was fine-tuned with the annotations of novel WM tracts using either the initialization strategy proposed in Sect.~\ref{sec:kt} or the more convenient implementation in Sect.~\ref{sec:warm}. 
	These two approaches are referred to as Ours1 and Ours2, respectively. 
	We compared our methodology with three competing methods. The first one is the baseline TractSeg network that was trained from scratch with the annotations of novel WM tracts. The second one is a representative conventional registration-based method Atlas FSL described in~\cite{wasserthal2019combined}, where an atlas is created from the available annotated scans and registered to test scans for segmentation. The third one is the classic fine-tuning method based on the segmentation model for existing WM tracts, which is referred to as FT.
	
	The proposed method was first evaluated qualitatively. Cross-sectional views of the segmentation results for representative test subjects and novel WM tracts are shown in Fig.~\ref{fig:qualitative}. The manual delineations are also shown for reference. It can be seen that the results of both of our strategies better resemble the manual delineations than the competing methods.
	
	Next, we evaluated the proposed method quantitatively. The Dice coefficient between the segmentation result and manual delineation was used as the evaluation metric. 
	For reference, we also computed the~\textit{upper bound}~(UB) performance, where abundant annotations were available for the novel WM tracts and the segmentation network was trained from scratch using these annotated scans. Specifically, the annotations of novel WM tracts on the 65 dMRI scans for training the network for existing WM tracts were also given, and they were used in conjunction with the other four scans annotated for the novel WM tracts to train a network that segments the novel WM tracts. 
	
	We computed the average Dice coefficient for each tract, and the results are summarized in Fig.~\ref{fig:boxplot_mean}. In addition, we compared Ours2 (which has slightly higher average Dice coefficients than Ours1) with the other methods (including Ours1) using paired Student's $t$-tests and measured the effect sizes (Cohen's $d$). The results are also shown in Fig.~\ref{fig:boxplot_mean}. We can see that our method (either Ours1 or Ours2) achieved higher Dice coefficients than the baseline, Atlas FSL, and FT, and the performance of our method is much closer to the upper bound than those of the competing methods. Also, Ours2 highly significantly $(p<0.001)$ outperforms the baseline method, Atlas FSL, and FT with large effect sizes ($d>0.8$). The performances of Ours1 and Ours2 are quite similar, which is indicated by the close average Dice coefficients and small effect size. Combined with the significant difference between them, these results show that Ours2 is consistently better than Ours1 with a very small margin, and this is consistent with our derivation and expectation in Sect.~\ref{sec:warm}.
	
	\begin{table}[!t]
		\caption{The means of the average Dice coefficients of the 12 novel WM tracts achieved with different numbers of annotated training scans. Our results are highlighted in bold. The effect sizes (Cohen's $d$) between Ours2 and the other methods are also listed. Asterisks indicate that the difference between Ours2 and the other method is significant using a paired Student's $t$-test. ($^{**}p<0.01$, $^{***}p<0.001$, n.s. $p\geq0.05$.)}
		\label{tab:number_dice}
		\centering
		\resizebox{0.78\columnwidth}{!}{
			
			\begin{tabular}{c c p{1.3cm}<{\centering} p{1.3cm}<{\centering} p{1.3cm}<{\centering} p{1.3cm}<{\centering} p{1.3cm}<{\centering} p{1.3cm}<{\centering} p{1.3cm}<{\centering}}
				\\
				\hline
				\hline
				Annotated  &  &  \multirow{2}{*}{Baseline} & \multirow{2}{*}{Atlas FSL} & \multirow{2}{*}{FT} & \multirow{2}{*}{Ours1} & \multirow{2}{*}{Ours2} & \multirow{2}{*}{UB}\\
				training scans & & & & & & & \\
				\hline
				\multirow{3}{*}{1} & Dice  & 0 & 0.645 & 0.590  &{\bf{0.777}} & {\bf{0.784}} & 0.828 \\
				& $d$ & 20.444 & 1.919 & 1.944  & 0.131 & - & - \\
				& $p$ & *** & *** & *** & ** & - & - \\
				\hline
				\multirow{3}{*}{5} & Dice  & 0.052 & 0.683 & 0.757  & {\bf{0.811}} & {\bf{0.812}} & 0.830 \\
				& $d$ & 10.362 & 2.004 & 1.000  & 0.021 & - & - \\
				& $p$ & *** & *** & *** & n.s. & - & - \\
				\hline
				\hline    
				
			\end{tabular}
		}
	\end{table}
	
	\subsection{Impact of the Number of Annotated Training Scans}
	\label{sec:number}
	In addition to the experimental setting above, to investigate the impact of the number of training scans annotated for the novel WM tracts on the segmentation accuracy, we considered two additional experimental settings, where additional annotated dMRI scans were included in the training and validation sets, or some of the annotated dMRI scans in the training and validation sets were excluded. Specifically, in the two cases, the numbers of annotated scans in the training/validation set were 1/0 and 5/2, respectively. The first case corresponds to a one-shot setting, and in this case model selection was performed based on the training data. The test set was not changed in these cases.
	
	We computed the average Dice coefficient for each tract and each method under these two settings, and the means of the average Dice coefficients are shown in Table~\ref{tab:number_dice}. 
	Here, we compared Ours2 with the other methods using paired Student's $t$-tests and measured the effect sizes. The UB performance was also computed. 
	Under these two settings, our method (either Ours1 or Ours2) still outperforms the competing methods. In particular, the performance of Ours2 is very close to the UB with five annotated training scans, and with only one annotated training scan for the novel WM tracts, the performance of either Ours1 or Ours2 is better than that of classic fine-tuning with five annotated training scans. In addition, Ours2 highly significantly outperforms the baseline, Atlas FSL, and FT with large effect sizes, and it is slightly better than Ours1.
	
	\begin{table}[!t]
		\caption{The means of the average RVDs of the 12 novel WM tracts. Our results are highlighted in bold. The effect sizes (Cohen's $d$) between Ours2 and the other methods are also listed. Asterisks indicate that the difference between Ours2 and the other method is significant using a paired Student's $t$-test. ($^{*}p<0.05$, $^{**}p<0.01$, $^{***}p<0.001$, n.s. $p\geq0.05$.)}
		
		\label{tab:rvd}
		\centering
		\resizebox{0.78\columnwidth}{!}{
			
			\begin{tabular}{c c p{1.3cm}<{\centering} p{1.3cm}<{\centering} p{1.3cm}<{\centering} p{1.3cm}<{\centering} p{1.3cm}<{\centering} p{1.3cm}<{\centering} p{1.3cm}<{\centering}}
				\\
				\hline
				\hline
				Annotated  &  &  \multirow{2}{*}{Baseline} & \multirow{2}{*}{Atlas FSL} & \multirow{2}{*}{FT} & \multirow{2}{*}{Ours1} & \multirow{2}{*}{Ours2} & \multirow{2}{*}{UB}\\
				training scans & & & & & & & \\
				\hline
				\multirow{3}{*}{1} & RVD  & 1 & 0.182 & 0.392  & {\bf{0.156}} & {\bf{0.151}} & 0.105 \\
				& $d$ & 17.458   &  0.403  &  1.854   & 0.067   & - & - \\
				& $p$ &  ***  &  *  & ***   &  n.s.  & - & - \\
				\hline
				\multirow{3}{*}{3} & RVD  & 0.986 & 0.175 & 0.207  & {\bf{0.130}} & {\bf{0.129}} & 0.107 \\
				& $d$ &  20.566  &  0.671  &   0.901  &  0.009  & - & - \\
				& $p$ & ***   &  *  &  *  &  n.s.  & - & - \\
				\hline
				\multirow{3}{*}{5} & RVD  & 0.955 & 0.199 & 0.158  & {\bf{0.129}} & {\bf{0.131}} & 0.105 \\
				& $d$ &  11.234  &  0.815  &  0.372   &  0.036  & - & - \\
				& $p$ & ***   &  *  &  *  &  n.s.  & - & - \\
				\hline
				\hline    
				
			\end{tabular}
		}
	\end{table}
	
	\subsection{Evaluation of Volume Difference}
	In addition to segmentation accuracy, we considered an additional evaluation metric, the \textit{relative volume difference}~(RVD) between the segmented novel WM tracts and the manual delineations, for all the experimental settings in Sects.~\ref{sec:accuracy} and \ref{sec:number}. This metric was considered because tract volume is an important biomarker~\cite{lebel2019review} that is often used to indicate structural alterations. A smaller RVD is desired as it indicates a smaller bias in the structural analysis. 
	
	For each experimental setting, where the number of scans annotated for the novel WM tracts in the training set was one, three, or five, we computed the average RVD for each novel WM tract, and the means of these average RVDs are reported in Table~\ref{tab:rvd}. Again, Ours2 was compared with the other methods using paired Student's $t$-tests and the effect sizes were computed. Also, the UB performance is listed for reference. Either Ours1 or Ours2 has better RVD values than the competing methods, and Ours2 is significantly better than these competing methods with mostly large ($d>0.8$) or medium ($d$ close to 0.5) effect sizes. The performances of Ours1 and Ours2 are still comparable, and they are much closer to the UB than the performances of the competing methods are.
	
	\section{Discussion}
	\label{sec:discussion}
	
	Classic fine-tuning discards the information in the task-specific layer of an existing model, whereas we propose to also incorporate the information in this task-specific layer during knowledge transfer for the segmentation of novel WM tracts. We have derived that in this way the task-specific layer for segmenting the novel WM tracts can be better initialized than the random initialization in classic fine-tuning. In addition, we have derived that the use of the information can be achieved more adaptively by inserting a warmup stage before classic fine-tuning. From a different perspective, this derivation also explains that warmup is beneficial to the transfer of knowledge about WM tracts because it implicitly allows a more comprehensive use of existing knowledge. 
	Our derivations are consistent with the experimental results under different settings.
	
	Our method may be extended to deep networks~\cite{zhang2020deep} that classify fiber streamlines as well. Those networks also comprise feature extraction and task-specific classification layers, and the knowledge transfer can incorporate the task-specific layers using the proposed method for classifying novel fiber streamlines. 
	
	\section{Conclusion}
	\label{sec:conclusion}
	We have explored the transfer of knowledge learned from the segmentation of existing WM tracts for few-shot segmentation of novel WM tracts. 
	Unlike classic fine-tuning, we seek to also exploit the information in the task-specific layer. The incorporation of this knowledge allows better initialization for the network that segments novel WM tracts. 
	Experimental results on the HCP dataset indicate the benefit of our method for the segmentation of novel WM tracts.

	\subsubsection{Acknowledgements}
	This work is supported by Beijing Natural Science Foundation (L192058 \& 7192108) and Beijing Institute of Technology Research Fund Program for Young Scholars. The HCP dataset was provided by the Human Connectome Project, WU-Minn Consortium and the McDonnell Center for Systems Neuroscience at Washington University.
	
	\bibliographystyle{splncs04}
	\bibliography{refs}
\end{document}